\documentclass[a4paper]{article}

\usepackage[english]{babel}
\usepackage[utf8x]{inputenc}
\usepackage[T1]{fontenc}

\usepackage[a4paper,top=3cm,bottom=2cm,left=3cm,right=3cm,marginparwidth=1.75cm]{geometry}

\usepackage{amsmath}
\usepackage{graphicx}
\usepackage[colorinlistoftodos]{todonotes}
\usepackage[colorlinks=true, allcolors=blue]{hyperref}

\title{A Challenge Set for French $\rightarrow$ English Machine Translation }
\author{Pierre Isabelle and Roland Kuhn \\
  National Research Council Canada\\
  {\tt first.last@nrc-cnrc.gc.ca} }

\begin{document}
\maketitle

\begin{abstract}
We present a \emph{challenge set} for French $\rightarrow$ English machine translation based on the  approach introduced in \cite{Isabelle2017}.  Such challenge sets are made up of sentences that are expected to be relatively difficult for machines to translate correctly because their most straightforward translations tend to be linguistically divergent. We present here a set of 506 manually constructed French sentences,  307 of which are targeted to the same kinds of structural divergences as in the paper mentioned above. The remaining 199 sentences are designed to test the ability of the systems to correctly translate difficult grammatical words such as prepositions.  We report on the results of using this challenge set for testing two different systems, namely Google Translate and DEEPL, each on two different dates (October 2017 and January 2018). All the resulting data are made publicly available. 
\end{abstract}

\section{Introduction}

Isabelle, Cherry and Foster\cite{Isabelle2017} introduce a \emph{challenge set} approach to evaluating machine translation (MT) systems. This approach is not meant as a substitute for traditional evaluation methods such as average BLEU or human scores on a held out portion of the training corpus. It is rather meant to supplement these methods with tools that directly measure the extent to which MT systems manage to tackle some of the more difficult translation problems. Thus, unlike traditional metrics, challenge sets provide developers with a fine-grained view of the remaining obstacles.

Ideally, one would like challenge sets to be constructed automatically. This is all the more desirable in that such sets are intrinsically language-pair dependent. But until automatic construction methods become available, we can turn to human experts for developing limited sets of challenging sentences. This is what \cite{Isabelle2017} did for English$\rightarrow$French machine translation (MT). However, challenge sets are not only language-pair dependent: they are also direction dependent.  For example, in English$\rightarrow$French translation there is a need to choose between the French verbs \emph{savoir} and \emph{connaître} as the correct translation for the English verb \emph{to know}. As it turns out, this depends on the syntactic nature of the complement of the verb. But in the opposite direction this problem does not arise: both \emph{savoir} and \emph{connaître} simply translate as \emph{to know}. This kind of asymmetry led us to develop a new challenge set that specifically targets French $\rightarrow$ English MT. 

In section 2, we describe the makeup of our new challenge set. In section 3 we report on the results of subjecting both Google Translate and DEEPL to the resulting challenge on two different dates: 5 October 2017 and 16 January 2018. As we will see, this constitutes an interesting way to track the systems' evolution.

\section{Makeup of the New Challenge Set}

In developing our French $\rightarrow$ English challenge set we closely followed the practices described in \cite{Isabelle2017}. In particular:

\begin{itemize}
\setlength\itemsep{-1pt}
\item We used short sentences that are each meant to bring into focus a single linguistic issue.  
\item All sentences are based on "common general vocabulary" because our goal is not to test lexical coverage but rather the system's ability to bridge specific linguistic divergences. 
\item We provide one reference translation for each sentence, notwithstanding the fact that other acceptable translations are usually possible.
\item Each sentence is accompanied with a yes-no question that focuses the attention of evaluators on the particular issue that the sentence is intended to test.
\item The evaluator's responses to each yes/no question completely determine the outcome of the evaluation or the relevant sentence. Consequently, translation errors that lie outside the scope of the yes-no questions will be ignored as irrelevant.
\item We use the same major classes of structural divergences as in the paper mentioned above: morpho-syntactic, lexico-syntactic and purely syntactic divergences. Each class is further subdivided into a set of subclasses which has a large overlap with those presented in the same paper.
\end{itemize}
One important difference with the earlier paper is that, in addition to testing the system's ability to deal with structural divergences, this new dataset includes some 199 examples that are intended to probe the systems' ability to cope with the notoriously difficult translation of grammatical words. This is achieved using groups of examples in which the same French grammatical word needs to be rendered in  different ways in the English translation.

In the remainder of this section we describe and illustrate the classes of linguistic difficulties that are built into our challenge set.

\subsection{Morpho-Syntactic divergences}

We use the term \emph{morpho-syntactic divergences} to refer to cases where the two languages differ in which grammatical features are overtly marked in the   morphology of corresponding words. Whenever a target language word requires a feature marking that is not explicitly marked in its source, the MT system needs to infer the relevant feature from the context. Our challenge set is probing that capability for the following cases:

\begin{itemize}

\item \textbf{Proclitic pronouns.} French complement pronouns often need to be procliticized, that is, moved to the left of the verb and phonetically attached to it. When translated into English, these pronouns need to be repositioned in their normal complement position. Moreover, the French clitic frequently underdetermines the grammatical features of the complement, such its gender, person and number and case. In the following example, the French clitic \emph{se} is not marked for gender but it needs to be translated as a neutral reflexive pronoun \emph{itself} because it co-refers with the neutral noun \emph{machine}. 

\begin{quote}
Cette machine \emph{se} répare elle-même.  $\rightarrow$ This machine repairs \emph{itself}.
\end{quote}

The following two examples illustrate that the French clitic \emph{leur} underdetermines the case/preposition marking of the corresponding English complement:
\begin{quote}

Je \emph{leur} ai parlé.  $\rightarrow$ I spoke \emph{to them}.
\end{quote}

\begin{quote}
Je \emph{leur} ai emprunté un livre.  $\rightarrow$ I borrowed a book \emph{from them}.
\end{quote}

\item \textbf{Chez soi.} The French form \emph{chez moi/toi/lui...} can translate as \emph{at home} but only when it is being used reflexively:

\begin{quote}
Mon fils est demeuré \emph{chez lui}.  $\rightarrow$ My son stayed $\left\{at\  his\  place\ |\ at\ home\right\}$.
\end{quote}

\begin{quote}
Ma fille est demeurée chez lui.  $\rightarrow$ My daughter stayed $\{$\emph{at his place} | \emph{*at home}$\}$. \footnote{We use the asterisk to mark a  translation as incorrect.}
\end{quote}

\item \textbf{Verb tense.} The overt French verb tense marking frequently underdetermines its English counterpart. In the following example, the French verb form is compatible with both the indicative and subjunctive mood while its English counterpart is explicitly marked as subjunctive. As a result the MT system must be able to determine whether or not the context is triggering the subjunctive mood:

\begin{quote}
Il est essentiel qu'il \emph{arrive} à temps.  $\rightarrow$ It is essential that he \emph{arrive} on time.
\end{quote}

\item \textbf{Verb tense concordance.} In sentences expressing two events with a specific time dependency, French and English often feature different tense concordance constraints. In the following example both verbs are semantically future but English, contrary to French, requires the subordinate verb to be in the grammatical present tense:

\begin{quote}
Max \emph{partira} dès que tu te \emph{lèveras}.
  $\rightarrow$ Max \emph{will leave} as soon as you $\{$\emph{*will get up} | \emph{get up}$\}$.
 \end{quote}

\end{itemize}

\subsection{Lexico-syntactic divergences}

We now turn to \emph{lexico-syntactic divergences}. We place under that heading all cases where the corresponding governing words of the two languages happen to organize their respective dependents in different ways. As a result, whenever such a governing word gets translated the system must be able to reorganize its dependents accordingly.

\begin{itemize}

\item \textbf{Argument switch.} In some cases, the most straightforward translation of a given verb requires a change in the order of the verb's arguments. This is the case when the French verb \emph{manquer à} is translated as \emph{to miss}:

\begin{quote}
\emph{Mary} manque beaucoup à \emph{John}. $\rightarrow$ \emph{John} misses \emph{Mary} a lot.

\end{quote}

\item \textbf{Manner of movement verbs.} In English, a completed movement is often expressed by using a verb that expresses a manner of moving (\emph{walk, climb, swim, etc.}) and combining it with a prepositional phrase that expresses the endpoint of the movement. In French, this is normally expressed by a more generic movement verb together with an adverbial expressing the manner of that movement. Here are two examples:
\begin{quote}
John aimerait \emph{traverser} l'océan \emph{à la nage}.  $\rightarrow$ John would like to \emph{swim across} the ocean.
\end{quote}
\begin{quote}
John \emph{entra dans} la salle \emph{en courant}.  $\rightarrow$ John \emph{ran into} the room.
\end{quote}

\item \textbf{Verb/adverb transposition.} Some French verbs tend to be expressed as adverbs in English. This involves a reorganization of the sentence in which a verb which is subordinate in French becomes the main verb in English.

\begin{quote}
Max a \emph{fini par comprendre} la difficulté. $\rightarrow$ Max \emph{finally understood} the difficulty.
\end{quote}

\item \textbf{Non-finite to finite clause.} It is quite common for a non-finite clause of French to be translated as a finite clause in English. This raises the difficulty of introducing an adequate subject as well as an adequate verb tense in the English translation.

\begin{quote}
Max \emph{croit connaître} la vérité. $\rightarrow$ Max \emph{thinks he knows} the truth.
\end{quote}

\begin{quote}
Mary \emph{croyait connaître} la vérité. $\rightarrow$ Mary \emph{thought she knew} the truth.
\end{quote}

\begin{quote}
Aussitôt \emph{son travail terminé}, Mary partit.
$\rightarrow$ As soon as \emph{her work was over}, Mary left.
\end{quote}

\item \textbf{\emph{"fact"} insertion}.
\begin{quote}
Cela \emph{provient de ce que} Max a trop dormi. $\rightarrow$ That \emph{arises from the fact that} Max has slept too much.
\end{quote}

\item \textbf{\emph{"how"} insertion}.
\begin{quote}
Max \emph{sait réparer} une cafetière. $\rightarrow$ Max \emph{knows how to repair} a coffee maker.
\end{quote}

\item \textbf{Middle voice.} The so-called middle-voice of French involves a pseudo-pronominal verb form whose interpretation is related to that of a passive sentence, often with a generic interpretation. It is most often translated in English with a passive form.
\begin{quote}
Ce type de moteur \emph{se répare} facilement. $\rightarrow$ This type of engine \emph{can be repaired} easily.
\end{quote}

\item \textbf{Control verbs.} So called \emph{subject-control verbs} take an infinitival complement whose subject is understood to co-refer with the subject of the control verb. In contrast, the understood subject of \emph{object-control verbs} is the object of the control verb. This difference can be brought to light when the infinitival complement is reflexive.

\begin{quote}
Max a convaincu sa fille de \emph{se} sacrifier. $\rightarrow$ Max convinced his daughter to sacrifice $\{$\emph{*himself}\ |\ \emph{herself}$\}$.
\end{quote}

\begin{quote}
Max a promis à sa fille de ne pas \emph{se} sacrifier. $\rightarrow$ Max promised his daughter not to sacrifice $\{$\emph{himself} | \emph{*herself}$\}$.
\end{quote}

\item \textbf{Mass versus count nouns.} Both languages make a grammatical distinction between count nouns (e.g. \emph{book}, \emph{table}, \emph{idea}) versus mass nouns (e.g. \emph{wine}, \emph{butter}, \emph{fear}). However, there are cases where a French noun and its English counterpart happen to fall on different sides of the divide. In such cases, a partitive noun may need to be introduced or deleted in the translation.

\begin{quote}
Max lui a donné \emph{un conseil}. $\rightarrow$ Max gave him $\{$\emph{*an advice} | \emph{a piece of advice}$\}$.
\end{quote}

\item \textbf{Factitives.} Some French verbs require the use of the auxiliary verb "faire" in order to receive an agentive reading. In such cases, that auxiliary must disappear in the translation.

\begin{quote}
Max a $\{$*\emph{fondu} | \emph{fait fondre}$\}$ la glace. $\rightarrow$ Max melted the ice.
\end{quote}

\begin{quote}
Max a $\{$*\emph{explosé} | \emph{fait exploser}$\}$ un rocher. $\rightarrow$ Max blew up a rock.
\end{quote}

\item \textbf{Two-position adjectives.} The correct translation of several French adjectives depends on whether they are placed before or after the noun they modify.

\begin{quote}
Une idée \emph{simple} n'est pas forcément mauvaise. $\rightarrow$ A $\{$\emph{simple} | \emph{*mere}$\}$ idea is not necessarily bad.
\end{quote}

\begin{quote}
La \emph{simple} idée de partir la terrorisait. $\rightarrow$ The $\{$\emph{*simple} | \emph{mere}$\}$ idea of leaving terrorized her.
\end{quote}

\item \textbf{Genitive.} Contrary to French, the preferred way to express genitives in English is not to use a prepositional phrase but rather the case marking \emph{'s}.

\begin{quote}
Il a pris le livre \emph{de mon frère aîné}.
 $\rightarrow$ He took \emph{my elder brother's} book.
\end{quote}

\end{itemize}

\subsection{Purely syntactic divergences}
The third type of divergence considered stems from the fact that some syntactic constructions only exist in one or the other language. Whenever a French sentence contains a construction that has no direct counterpart in English, the MT system needs to be able to recast the source language material into a different construction. 

In fact, we have already seen one such case above, namely that of French proclitics. While we listed them  under the heading of morpho-syntactic divergences, they do exemplify both types of divergence at once. Since there are no proclitics in English, a French object proclitic needs to be relocated in the standard post-verbal position  in the English translation: \emph{Il la voit.} $\rightarrow$ \emph{He sees her}. Here are some  other subtypes of purely syntactic divergences.

\begin{itemize}
\item \textbf{Yes-no question syntax.} French and English differ in the way yes-no questions are formed. Basically, French questions are obtained as follows: if the subject is a proclitic, move it after the verb; otherwise insert a particle (either \emph{est-ce que} at the beginning of the sentence or \emph{-il} after the verb). In contrast, English questions are obtained by fronting an auxiliary verb.
\begin{quote}
As\emph{-tu} lu ce livre?
 $\rightarrow$ \emph{Have} you read this book?
\end{quote}
\begin{quote}
Max partira\emph{-t-il} à temps?
 $\rightarrow$ \emph{Will} Max leave on time?
\end{quote}

\item \textbf{Tag questions.} The so-called \emph{tag-question} construction of English does not exist in French, but the French \emph{n'est-ce pas?} sentence-final question is normally translated as an appropriate tag question, which involves selecting the right auxiliary verb.
\begin{quote}
Il a vu la photo hier, \emph{n'est-ce pas}? 
 $\rightarrow$ He saw the picture yesterday, \emph{didn't he?}
\end{quote}
\begin{quote}
Nous devrions vérifier le niveau d'huile, \emph{n'est-ce pas}?
 $\rightarrow$ We should check the oil level, \emph{shouldn't we}?
\end{quote}

\item \textbf{WH-movement: relative clauses.} When a relative clause is formed, its internal relativized element gets fronted, typically in the form of a "WH-word". For example, in \emph{The man whom you saw is my brother} the word "whom" is understood to refer to the object of the verb "saw", which we will call its \emph{native site}. French and English relative constructions are often parallel enough that an MT system can get away with a superficial process that falls short of explicitly relating the WH word to its native site. However, such a superficial approach breaks down in the case of \textbf{stranded prepositions}. In French, whenever a prepositional phrase is relativized, its preposition must be fronted alongside the WH-word: \emph{la fille \textbf{avec qui} tu as dansé}. In contrast, English will often leave the preposition stranded: \emph{the girl you danced \textbf{with}}. Note that in the French $\rightarrow$ English direction, the MT system does not have to move the preposition to its native site, since preposition fronting is also permitted in English. However, if the system does move the preposition to its correct native site, then this provides nice evidence that it is able to perform some deeper processing.

\begin{quote}
L'homme \emph{à qui} Max a donné un livre est parti.  $\rightarrow$ The man \emph{whom} Max gave a book \emph{to} is gone.
\end{quote}
\begin{quote}
La fille \emph{dont} il a parlé est brillante.
 $\rightarrow$ The girl \emph{that} he talked \emph{about} is brilliant.
\end{quote}

\item \textbf{WH-movement: interrogatives.} Question formation and relative formation are highly parallel in both French and English. As a result, stranded prepositions raise the same translation issues with questions as with relatives.  

\begin{quote}
\emph{À qui} Max a-t-il donné un livre?   $\rightarrow$ \emph{Whom} did Max gave a book \emph{to}?
\end{quote}

\begin{quote}
\emph{Pour quelle} compagnie travaille-t-il?
 $\rightarrow$ \emph{What} company does he work \emph{for}?
\end{quote}

\item \textbf{Negation.} In French, negation is typically expressed using a discontinuous form such as \emph{ne ... pas/jamais/plus/nullement} while in English this is typically done using a single word. MT systems often run into difficulty with this phenomenon. In our first example below the system needs to recognize that \emph{ne} is being used in an "expletive" (i.e. non-negative) way and therefore should not be translated.  In our second example, the French negation is to be rendered by the single negation word \emph{not}, but while reinforcing it with the intensifying adverbial \emph{at all}.

\begin{quote}
Je crains que Max \emph{ne} vienne nous voir. $\rightarrow$ I'm afraid Max is coming to see us.
\end{quote}

\begin{quote}
Max \emph{ne} comprend \emph{nullement} cette idée. $\rightarrow$ Max does \emph{not} understand this idea \emph{at all}.
\end{quote}

\item \textbf{Double negation.} Double negations are sometimes used for stylistical effect and some MT systems appear to have difficulty coping with that.

\begin{quote}
Ce politicien n'est pas capable de ne pas mentir. $\rightarrow$ This politician is not able not to lie.
\end{quote}

\begin{quote}
C'est le docteur dont il est  \emph{impossible} que vous \emph{n'}ayez \emph{pas} entendu parler. $\rightarrow$ It is the doctor of whom it is \emph{impossible} that you have \emph{not} heard.
\end{quote}

\item \textbf{Other doubled concepts.} Some MT systems appear to experience some difficulty with sentences that contain two tokens for the same concept.

\begin{quote}
Il a commis \emph{faute} sur \emph{faute}. $\rightarrow$ He committed \emph{mistake} after \emph{mistake}.
\end{quote}

\begin{quote}
C'est \emph{beaucoup beaucoup} mieux. $\rightarrow$ This is \emph{much much} better.

\end{quote}

\end{itemize}

\subsection{Purely lexical divergences.}
The kinds of structural divergences described above closely mirror what was done in \cite{Isabelle2017} for English$\rightarrow$French machine translation. However, in that work idiomatic phrases and support verbs were placed under the broader category of lexico-syntactic divergences. In the present work, we instead introduce an additional top-level category, namely that of purely lexical divergences. Alongside testing material for idioms and support verbs, this category will include a substantial amount of additional material meant to test the ability of MT systems to translate common grammatical words such as prepositions. 

\begin{itemize}

\item \textbf{Common idioms -- fixed.} Some phrases need to be translated as a group because they happen to have a language-specific idiomatic meaning. The simpler case is that of fixed idioms, those that always appear under one and the same form. 

\begin{quote}
Ils sont déterminés à continuer \emph{envers et contre tous}. $\rightarrow$ They are determined to continue \emph{in spite of all opposition}.
\end{quote}

\begin{quote}
Ils partiront \emph{entre chien et loup}. $\rightarrow$ They will leave \emph{at dusk}.
\end{quote}

\item \textbf{Common idioms -- variable.} Many idioms exhibit some morphological and/or syntactic flexibility. As a result, there is a need for MT systems to generalize over a range of different surface forms.

\begin{quote}
Cessez de \emph{tourner autour du pot}. $\rightarrow$ Stop \emph{beating around the bush}.
\end{quote}

\begin{quote}
Il \emph{tournait} constamment \emph{autour du pot}. $\rightarrow$ He was constantly \emph{beating around the bush}.
\end{quote}

\begin{quote}
Vous \emph{mettez la charrue devant les boeufs}.
$\rightarrow$ You \emph{put the cart before the horse}.
\end{quote}

\begin{quote}
La \emph{charrue} a été \emph{mise avant les bœufs}. $\rightarrow$ The \emph{cart} was \emph{put before the horse}.
\end{quote}

\item \textbf{Support verbs.} These verbs (also known as "light verbs") carry little meaning in themselves. Rather they combine with their complement to express what can often be expressed as a single verb. For example, \emph{to walk} and \emph{to take a walk} are roughly equivalent. But even though the support verb - here, \emph{take} - carries little meaning in itself, its choice is not free. In this example, \emph{*make a walk} is not an acceptable substitute. Support verbs must be translated as a whole with their complements. 

\begin{quote}
Max a \emph{fait campagne contre} le maire hier.
 $\rightarrow$ Max \emph{campaigned against} the mayor yesterday.
\end{quote}

\begin{quote}
Ceci \emph{apporte la preuve qu}'il était au courant. $\rightarrow$ This \emph{is proof that} he was aware.
\end{quote}

Unacceptable, literal translations for these two examples would be:
\begin{quote}
 Max \emph{*made a campaign against} the mayor yesterday.
\end{quote}

\begin{quote}
 This \emph{*brings proof that} he was aware.
\end{quote}

\item \textbf{Grammatical words.} Grammatical words such as prepositions are notoriously difficult to translate. Our challenge set includes testing material for some 28 different grammatical words or phrases that are relatively difficult to translate correctly because they each have multiple uses. For each one we provide sets of sentences where the word needs to receive different translations as a result of these different uses. Consider for example some different uses/translations of the French preposition \emph{en}:

\begin{quote}
Il lui a offert un foulard \emph{en} soie.  $\rightarrow$ He offered her a silk scarf.
\end{quote}

\begin{quote}
Il est docteur \emph{en} philosophie. $\rightarrow$ He's a doctor \emph{of} philosophy.
\end{quote}

\begin{quote}
\emph{En} semaine, je travaille.  $\rightarrow$ \emph{On} weekdays, I work.
\end{quote}

\begin{quote} 
J'ai payé mes études \emph{en} vendant du café. $\rightarrow$ I paid my tuition \emph{by} selling coffee.
\end{quote}

\begin{quote} 
\emph{En} travaillant, j'aime écouter de la musique.  $\rightarrow$ \emph{While} working, I like to listen to music.
\end{quote}

Another good example is the multiple uses and translations of the preposition \emph{par}:

\begin{quote} 
Il a été averti \emph{par} Paul.  $\rightarrow$ He was warned \emph{by} Paul.
\end{quote}

\begin{quote} 
Un lundi \emph{par} mois, il se rend au marché.  $\rightarrow$ One Monday \emph{per} month, he goes to the market.
\end{quote}

\begin{quote} 
Il a fait cela \emph{par} plaisir.  $\rightarrow$ He did it \emph{for} pleasure.
\end{quote}

\begin{quote} 
Il a fait cela \emph{par} habitude.  $\rightarrow$ He did it \emph{out of} habit.
\end{quote}

\begin{quote} 
Le bateau a coulé \emph{par} cent mètres de fond.  $\rightarrow$ The boat sank \emph{to} a depth of a hundred meters. 
\end{quote}

\end{itemize}

\subsection{Our New Challenge Set.}

We manually developed a set of 506 different challenging examples populating the main categories discussed above with the distribution shown in Table 1.

\begin{table}[b]
  \begin{center}
  \begin{tabular}{lrrr}
    \hline
    Category & No. of examples & Percent \\
    \hline
    Morpho-synctactic & 43 & 8.5\% \\
    Lexico-syntactic & 79 & 15.6\%  \\
    Purely syntactic & 84 & 16.6\% \\
    Purely lexical & 300 & 59.3\% \\
    \hline
    Total & 506 & 100\%
  \end{tabular}
  \caption{Distribution of challenge set examples across main categories.}
  \label{tab:data}
  \end{center}
\end{table}

In addition to making use of our own personal experience in machine translation, we were able to draw many examples from Morris Salkoff's highly detailed and insighful French-English contrative grammar \cite{Salkoff:99}.

\section{Testing Google Translate and DEEPL}

Armed with this new French $\rightarrow$ English challenge set, we decided to evaluate the performance of the Google Translate and DEEPL neural machine translation systems. We submitted all 506 sentences to each system on two different dates: 5 October 2017 and 16 January 2018. We collected the results and proceeded to evaluate them.

The evaluation protocol was as follows. The human evaluator looks at each test case in turn, being provided with: a) the source-language sentence; b) one reference translation; c) the machine-translated sentence to be evaluated; and d) a single yes/no question about the translation and its relationship to the source-language sentence. The evaluator simply provides an answer the yes/no question associated with each translated example. Figure 1 provides two examples of material being presented to the evaluator together with his/her response (either "Yes" or "No").

\begin{figure}
\begin{tabular}{|lp{0.8\columnwidth}|}
\hline
Src & La femme \textbf{s'}est regardée dans le miroir.
 \\
Ref & The woman looked at herself in the mirror.
 \\
Sys & The woman looked at herself in the mirror.
\\
\hline
\multicolumn{2}{|l|}{Is the French highlighted pronoun correctly translated (y/n)?
 \textbf{Yes}}\\
\hline
\end{tabular}

\vspace{5pt}

\begin{tabular}{|lp{0.8\columnwidth}|}
\hline
Src & Je \textbf{le} suppose.
 \\
Ref & I suppose so.
 \\
Sys & I'm guessing.
\\
\hline
\multicolumn{2}{|l|}{Is the French highlighted pronoun correctly translated (y/n)?
 \textbf{No}}\\
\hline
\end{tabular}
\caption{Example challenge set questions.\label{fig:example1}}
\vspace{-5pt}
\end{figure}

The first author made an initial pass at responding to each one of the 2024 relevant questions  (506 for each one of the four machine translations). The second author checked all these judgments and noted all disagreements. Each difference was then discussed by the two authors and a joint decision was made. 

Thus, unlike in \cite{Isabelle2017} where three independent evaluators were used, the results presented below only rely on the authors' judgments. However, we are making these judgments available alongside the new challenge set so that interested parties can compare them with their own judgments.

The main results are presented in Table 2. The outcome of October 2017 was similar to that presented in \cite{Isabelle2017} for the English$\rightarrow$French direction: in both cases DEEPL turned out to deal with the challenge set quite a bit better that Google. On the present challenge set, DEEPL's overall rate of success was almost 13\% higher than that of Google. This advantage holds in all categories of examples except for morpho-syntax where both systems are tied.

We can also see that the overall performance of both systems turned out to be somewhat better in January 2017. The Google system achieved an overall improvement of 2.6\% for a relative error reduction of 3.6\%, while the DEEPL system got a 1\% improvement for a relative error reduction of 1.3\%. The rate of progress varied across categories. In the case of morpho-syntax, Google managed to gain 7\%,  significantly bettering DEEPL which turned out to lose 4.6\%. Conversely, in the case of pure syntax Google lost 3.5\% while DEEPL's performance remained unchanged. For the other two categories (purely lexical and lexico-syntactic) both systems progressed but Google did so more markedly.

Table 3 provides a breakdown of the same results in terms of our finer-grained subcategories.

\begin{table}[b]
  \begin{center}
  \begin{tabular}{lrrrrr}
    \hline
    Divergence type & GNMT-1 & GNMT-2 &  DEEPL-1 & DEEPL-2 \\
    \hline
    Morpho-syntactic & 76.7\%	& 83.7\%  & 76.7\%  & 72.1\%  \\
    Lexico-syntactic & 63.3\%	& 63.3\%  & 75.9\%	& 78.5\%  \\
    Purely Syntactic & 70.2\%     & 66.7\%	& 79.8\%  & 79.8\% \\\
    Purely Lexical   & 64.8\%     & 66.3\%  & 80.9\%  & 81.4\%  \\
    \hline
    Overall& 63.6\% &  66.2\% & 76.5\% & 77.5\% \\
    \hline
  \end{tabular}
  \caption{Challenge set success rate for the systems under study, with "-1" and "-2" indicating  respectively the versions of September 2017 and January 2018.}

  \label{tab:results}
  \end{center}
\end{table}

\begin{table*}[tb]
\centering
\begin{tabular}{llrrrrr}
\hline
Category         & Subcategory   & \# & GNMT-1   & GNMT-2     & DEEPL-1	& DEEPL-2\\
\hline
M-Syn & Proclitic pronouns  & 21  & 81.0\%   & 90.5\% & 95.2\% & 85.7\%\\
      & "Chez soi"  & 4  & 50.0\%  & 50.0\%  & 50.0\%  & 50.0\% \\
      & Verb tense  & 11  & 81.8\%   & 81.8\% & 72.7\%  & 72.7\%\\
      & Verb tense concord & 7 & 71.4\%  & 85.7\%  & 42.9\% & 42.9\% \\
\hline
L-Syn & Argument switch               & 5  & 40\%   & 20\%   & 80\%  &100\% \\
            & Manner-of-movement verbs      & 6  & 33.3\%   & 33.3\%   & 50\%   & 50\% \\
            & Verb-adverb transposition      & 3  & 66.7\%  & 66.7\%  &100\% & 66.7\%  \\
            & Non-finite $\rightarrow$ finite clause        & 20  & 80\%   & 80\%  & 75\%  &80\%\\
            & "De/à ce que"   $\rightarrow$ "from the fact that" & 2  & 0\%  & 50\%  & 50\%  & 50\%\\
            & V1 V2\textsubscript{inf} $\rightarrow$  V1 how to V2\textsubscript{inf} & 2  & 100\%  & 100\%   & 100\%  & 100\% \\
            & Middle voice & 3  & 66.7\%   & 100\%   & 66.7\% & 100\% \\
            & Control verbs      & 5  & 40\%   & 60\%   & 60\%   & 60\% \\
            & Count Vs mass nouns      & 6  & 83.3\%   & 83.3\%   & 66.7\%   & 66.7\% \\
            & "Voilà [TIME] que" & 3 & 66.7\% & 0\% & 100\% & 100\% \\
            & Factitives      & 13 & 38.5\%   & 38.5\%   & 76.9\%   & 69.2\% \\
            & Two-position adjectives      & 9 & 88.9\%   & 88.9\%   & 88.9\%   & 100\% \\ 
            & Genitives    & 2 & 100\%   & 100\%   & 100\%   & 100\% \\
\hline
Syn        & Yes-no question syntax   & 4  & 50\%  & 50\% & 100\% & 100\%\\
           & Tag questions            & 4  & 0\%   & 50\%   & 100\% & 100\%\\
           & WH movement, relatives   & 19  & 78.9\%   & 68.4\%   & 78.9\% & 78.9\%\\
           & WH movement, questions   & 10  & 80\%   & 70\%   & 90\%  & 90\%\\
           & Negation                 & 8  & 75\%   & 87.5\%   & 100\% & 100\%  \\
           & Double negation          & 20  & 55\%  & 50\%  & 65\%  & 65\%\\
           & Other doubled concepts   & 5  & 80\%  & 40\%  & 20\%  & 20\%\\
           & Inalienable possession   & 6  & 100\% & 100\% & 100\% & 100\%\\
           & Subject inversion        & 8  & 87.5\%  & 87.5\%  & 87.5\% & 87.5\% \\
\hline
Lex        & Common idioms -- fixed   & 25 & 32\%  & 40\%   & 52\%  & 48\%\\
           & Common idioms -- variable   & 24 & 33.3\%  & 58.3\%   & 66.7\%  & 66.7\%\\
           & Support verbs   & 52 & 67.3\%  & 71.2\%   & 71.2\%  & 80.8\%\\
           & Translation of grammatical words   & 199 & 64.8\%  & 66.3\%   & 80.9\%  & 81.4\%\\
           
\hline
\end{tabular}
\caption{
  Summary of scores by fine-grained categories. ``\#'' reports number of questions in each category,
  while the reported score is the percentage of questions for which the divergence was correctly bridged.}
\label{tab:fine}
\end{table*}

\subsection{Conclusion}

We have presented a new challenge set for evaluating machine translation systems in the French$\rightarrow$English direction based on the principles presented in \cite{Isabelle2017}. This new set includes 506 different sentences spread across four categories: morpho-syntactic, lexico-syntactic, purely syntactic and purely lexical. The first three categories mirror those of \cite{Isabelle2017} but the last one is novel. Each sentence is meant to test the ability of MT systems to bridge one specific divergence issue between the two languages.

Our 506 challenge sentences have been submitted to the Google and DEEPL MT systems on two different dates: 5 October 2017 and 16 January 2018. The results have been evaluated according to the method presented in \cite{Isabelle2017}, which amounts to responding to the yes-no questions attached to each challenge sentence. 

In this case the evaluators were the co-authors of this paper, which is not optimal. However, we are making all the data available so that readers can compare our judgments with theirs.

\bibliographystyle{plain}
\bibliography{challenge}

\end{document}